\documentclass{article}
\usepackage{spconf,amsmath,graphicx}
\usepackage{amssymb}
\usepackage{booktabs}
\usepackage[hidelinks]{hyperref}
\usepackage{xurl}

\usepackage{tikz}
\usetikzlibrary{
    arrows.meta,
    positioning,
    calc,
    shapes.geometric,
    fit
}

\tikzset{
    flow/.style={
        ->,
        thick,
        >=Latex
    }
}

\tikzset{
    residual/.style={
        ->,
        thick,
        dashed,
        orange!80!black,
        >=Latex
    }
}

\tikzset{
    condition/.style={
        draw,
        rounded corners,
        minimum width=2.5cm,
        minimum height=1.5cm,
        align=center,
        fill=green!10,
        thick
    }
}

\tikzset{
    masknet/.style={
        draw,
        rounded corners,
        minimum width=2.5cm,
        minimum height=1.4cm,
        align=center,
        fill=orange!12,
        thick
    }
}

\tikzset{
    controlnet/.style={
        draw,
        rounded corners,
        minimum width=2.7cm,
        minimum height=1.3cm,
        align=center,
        fill=orange!15,
        thick
    }
}

\tikzset{
    backbone/.style={
        draw,
        rounded corners,
        minimum width=3.6cm,
        minimum height=1.7cm,
        align=center,
        fill=blue!10,
        thick
    }
}

\tikzset{
    modality/.style={
        draw,
        rounded corners,
        minimum width=2.6cm,
        minimum height=1cm,
        align=center,
        fill=purple!8,
        thick
    }
}

\tikzset{
    output/.style={
        draw,
        rounded corners,
        minimum width=2.4cm,
        minimum height=1.4cm,
        align=center,
        fill=gray!5,
        thick
    }
}
\tikzset{%
imagebox/.style={%
    draw,
    rounded corners,
    minimum width=2.0cm,
    minimum height=1.6cm,
    align=center,
    fill=gray!4,
    thick
}%
}

\tikzset{%
smallimagebox/.style={%
    draw,
    rounded corners,
    minimum width=1.45cm,
    minimum height=1.05cm,
    align=center,
    fill=gray!4,
    thick,
    font=\scriptsize
}%
}

\tikzset{%
stage/.style={%
    font=\bfseries
}%
}

\usepackage{enumitem}
\setlist{nosep, leftmargin=14pt}

\usepackage{mwe} % to get dummy images

\def\x{{\mathbf x}}

\title{Spatially and Compositionally Controlled Multimodal Glioblastoma MRI Synthesis with Latent Diffusion}
\name{%
\begin{tabular}{@{}c@{}}
Steven Glass$^{\star}$ \qquad
Max Van Puyvelde$^{\dagger}$ \qquad
Marija Pizurica$^{\dagger}$ \\
Mia Meyer$^{\star}$ \qquad
Olivier Gevaert$^{\dagger}$ \qquad
Willie Brink$^{\star}$
\end{tabular}
}

\address{
$^{\star}$Department of Mathematical Sciences, Stellenbosch University \\
$^{\dagger}$Department of Computational Medicine, Stanford University
}

\begin{document}
%\ninept
%
\maketitle
\begin{abstract}
Glioblastoma exhibits substantial heterogeneity in tumour morphology and composition across MRI modalities, often making available clinical datasets highly unbalanced in terms of patient pathologies. We propose a two-stage latent diffusion model (LDM) framework for the generation of explicitly controllable 2D glioblastoma MRI scans. First, a segmentation LDM generates tumour maps conditioned on location, spatial extent, axial position, and subregion composition. Second, a modality-conditioned LDM with ControlNet generates T1, T1Gd, T2, and FLAIR images adhering to the requested segmentation. On a held-out test set, generated segmentations achieved a mean centroid error of 0.842 pixels and a subregion proportion error of 0.795 percentage points. ControlNet improved Fréchet inception distance (FID) across all four modalities, reducing pooled FID from 42.59 to 36.12. An expert reader study yielded mean quality ratings of 6.60 for real images, 6.83 for ControlNet-generated images, and 6.77 for fully synthetic images.
\end{abstract}


\section{Introduction}
\label{sec:intro}

% Biomed motivation

% Existing Work

% Research Gap

% What I contribute (tex diagram here)

Glioblastoma is an aggressive form of brain cancer with a high degree of heterogeneity across T1, T1Gd, T2 and FLAIR MRI. The diversity observed in finite clinical glioblastoma datasets is only partially representative of the true variability of real-world patients, limiting the robustness of data-driven models. Recent advances in diffusion models have demonstrated that generative models are able to produce high-fidelity images while capturing useful variation \cite{hoddpm},\cite{dhariwal2021diffusionmodelsbeatgans}. However, unconditional generation alone cannot deliberately oversample rare tumour locations, sizes, or subregion compositions.

We address this with a two-stage framework that separates pathology generation from MRI synthesis. Our main contributions are: (i) a conditional generative model (Seg-LDM) for explicit spatial and compositional control of synthetic tumour segmentation maps; (ii) a two-stage generative framework that propagates a requested pathology into the synthesis of multi-modal brain MRI scans; (iii) quantitative evaluation of condition adherence and image fidelity, complemented by an expert reader study.

\section{Related Work}

Denoising diffusion models \cite{hoddpm} have emerged as a powerful generative modelling approach. In medical imaging, Pinaya et al.\ \cite{pinaya2022brainimaginggenerationlatent} demonstrated that latent diffusion models (LDMs) are capable of generating high-resolution 3D brain MRI scans, while Khader et al.\ \cite{khader2023medicaldiffusiondenoisingdiffusion} demonstrated the utility of synthetic biomedical images on downstream tasks via data augmentation.

Recent work has addressed the challenge of introducing spatially conditioned generation. Med-DDPM \cite{dorjsembe2024conditional} conditions brain MRI generation using semantic masks, Kebaili et al.\ \cite{KEBAILI2025MultiModal} jointly generate multimodal brain MRI and tumour masks, and Zhang et al.\ \cite{zhang2024lefusion} demonstrated controllable pathology generation with LeFusion, which conditions lesion synthesis on spatial location, size, and boundary.

ControlNet \cite{zhang2023ControlNet} provides an architecture for introducing spatial guidance while preserving a pretrained diffusion backbone.
Most similar to our overall pipeline, Kim et al.\ \cite{kim20263dlldmlabelguided3dlatent} introduced 3D-LLDM  which first generates tumour segmentation maps which subsequently guide MRI generation using ControlNet.

\section{Method}

\begin{figure}[t]
    \centering
    \resizebox{\columnwidth}{!}{%
        \input{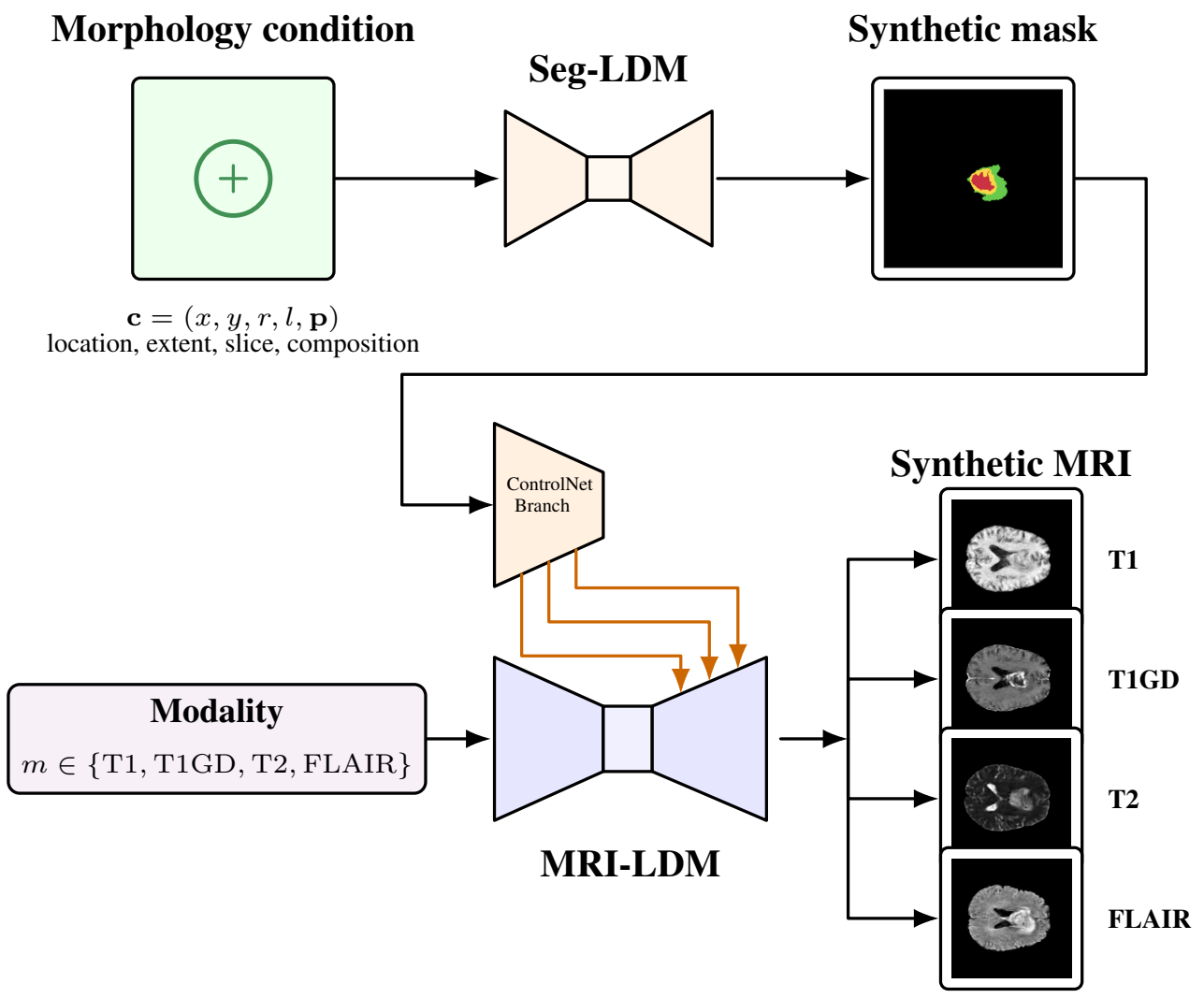}
    }
    \caption{The complete generative pipeline of our approach.}
\end{figure}

\subsection{Preliminaries}

First introduced by Rombach et al.\ \cite{rombach2022LDM}, LDMs are a variant of diffusion models that perform the diffusion process in a learned, lower-dimensional latent space instead of in image space, which reduces the computational complexity of both training and inference. Consider image data $\mathbf{x} \in \mathbb{R}^{C\times H\times W}$, where $C, H$ and $W$ represent the number of channels, height and width respectively. The encoder $\mathcal{E}_\phi$ of a variational autoencoder (VAE) maps $\mathbf{x}$ to a latent representation
$\mathbf{z}_0$, which is later decoded by the decoder $\mathcal{D}_{\psi}$. The forward diffusion process corrupts $\mathbf{z}_0$ according to
\begin{equation}
    \mathbf{z}_t = \sqrt{\bar{\alpha}_t} \mathbf{z}_0 + \sqrt{1 - \bar{\alpha}_t} \boldsymbol{\epsilon},
    \qquad
    \boldsymbol{\epsilon}\sim\mathcal{N}(\mathbf{0},\mathbf{I}),
\end{equation}
\noindent
and a conditional denoising network
$\boldsymbol{\epsilon}_\theta(\mathbf{z}_t,t,\mathbf{c})$ is trained to reverse the added noise at each timestep by minimising the loss
\begin{equation}
\label{eq. loss}
    \mathcal{L}_{\mathrm{diff}} = \mathbb{E}_{\mathbf{z}_0, \boldsymbol{\epsilon}, t, \mathbf{c}} \left[\left\|\boldsymbol{\epsilon} - \boldsymbol{\epsilon}_\theta (\mathbf{z}_t,t,\mathbf{c}) \right\|_2^2 \right].
\end{equation}
\noindent
Here $\mathbf{c}$ denotes the conditioning variables. At inference, the learned reverse process transforms Gaussian noise into a latent sample $\mathbf{z}_0$, which gets decoded back into an image $\mathbf{x}_0$.

% Data and Preprocessing
\subsection{Data and preprocessing}

We combine the UCSF-PDGM MRI ($n=494$) \cite{calabreseUCSFGlioblastoma2022} and the UPenn-GBM ($n=611$) \cite{bakasUPENNGlioblastoma2022} datasets, yielding $n=1105$ subjects with T1, T1Gd, T2, and FLAIR scans and corresponding tumour segmentations. Both cohorts were skull-stripped, and preprocessed using the BrainLes preprocessing package \cite{kofler2025brainles}: negative intensities were clipped, N4 bias-field correction was applied, modalities were co-registered to T1, and volumes were affinely transformed to the SRI24 atlas. Resulting volumes have $1\mathrm{mm}$ isotropic resolution and size $240\times240\times155$.

Tumour segmentation maps were transformed into the same space using nearest-neighbour interpolation. Subjects were partitioned at the patient level before axial slice extraction, and all slices were resized to $256\times256$. For MRI synthesis, subjects were randomly assigned to disjoint backbone (75\%) and ControlNet (25\%) partitions, each being split into 80/10/10 training, validation, and testing sets stratified by cohort. Both VAEs and the Seg-LDM used the unions of the corresponding training, validation and testing subsets. The MRI backbone used only the backbone subsets, while ControlNet used only the ControlNet subsets with the backbone frozen.

% Controllable Mask Gen
\subsection{Controllable tumour segmentation map generation}

A 2D VAE adapted from Van Puyvelde et al.\ \cite{puyvelde2026CONFLUX} is trained to encode a segmentation map $\mathbf{y}\in\{0,1\}^{4\times256\times256}$ onto a latent representation $\mathbf{z}_0\in\mathbb{R}^{4\times32\times32}$. After training, the VAE is frozen and a conditional LDM (Seg-LDM) with a U-Net architecture \cite{ronneberger2015unet} learns the distribution of the latent representations of the segmentation maps, conditioned on coarse spatial instruction and global composition. Let $\mathcal{S}$ denote the set of tumour foreground pixels. Its centroid $\boldsymbol{\mu}$ and radius $r$ are
\begin{equation}
    \boldsymbol{\mu}=\frac{1}{|\mathcal{S}|}\sum_{\mathbf{v}\in \mathcal{S}}\mathbf{v},\qquad r=\max_{\mathbf{v} \in \mathcal{S}}||\mathbf{v} - \boldsymbol{\mu}||_2.
\end{equation}
\noindent
These define the binary spatial disk $\mathbf{S}$, which specifies a coarse spatial region for the tumour, while allowing the model to synthesize the tumour morphology. This disk is downsampled and concatenated channel-wise with the noised latent $\mathbf{z}_t$. Global conditioning is represented by the vector
\begin{equation}
    \mathbf{g}=[l,p_1,p_2,p_3]^T ,
\end{equation}
\noindent
where $l\in[0,1]$ is the normalised axial slice index, and $p_k$ is the proportion of tumour region $k\in\{1,2,3\}$, which represent necrotic, edema, and enhancing regions respectively. Separate multilayer perceptrons embed $\mathbf{g}$ and diffusion timestep $t$ into a common embedding space, where they are added and supplied to the residual blocks of the denoising U-Net. Classifier-free guidance \cite{ho2022classifierfreediffusionguidance} is used during sampling to control adherence to the requested conditions. During training, all conditioning is randomly removed with $p_{\text{uncond}}=0.1$. The Seg-LDM is trained using Eq.\ \ref{eq. loss} with conditioning $(\mathbf{S}, \mathbf{g})$.

% Segmentation guided MRI synthesis
\subsection{Segmentation-guided MRI generation}

A separately trained VAE encodes the MRI slice $\mathbf{x}\in \mathbb{R}^{256\times256}$ onto the latent representation $\mathbf{z}_0\in \mathbb{R}^{16\times32\times32}$. The MRI-LDM is a U-Net LDM backbone trained on the latent representations of slices across all four modalities. A requested modality $m\in\{\mathrm{T1},\mathrm{T1Gd},\mathrm{T2},\mathrm{FLAIR}\}$ is embedded and added to the timestep embedding in the same manner as the Seg-LDM. During training, modality conditioning is randomly removed with $p_{\text{uncond}}=0.1$, allowing for classifier-free guidance. The MRI-LDM is trained using Eq. \ref{eq. loss} with conditioning $m$.

Once training of the backbone model is complete, the parameters are frozen, and spatial control is introduced via a ControlNet branch. MRI synthesis is guided according to a requested tumour segmentation map $\mathbf{y}$, which is first projected onto the same spatial resolution as the MRI latents. A trainable copy of the MRI-LDM's encoder processes $\mathbf{y}, \mathbf{z}_t$ and $t$. Multi-scale ControlNet features are then processed by zero-initialised $1\times1$ convolutional layers, and additively combined with the backbone skip features. As a result, the ControlNet branch learns how the requested tumour segmentation map should alter the denoising trajectory, producing MRI scans with a matching tumour pathology.

%TODO, mustnt forget to mention the optimiser and hyperparams

\section{Results and Discussion}

\begin{figure}
    \centering
    \includegraphics[width=1.0\linewidth]{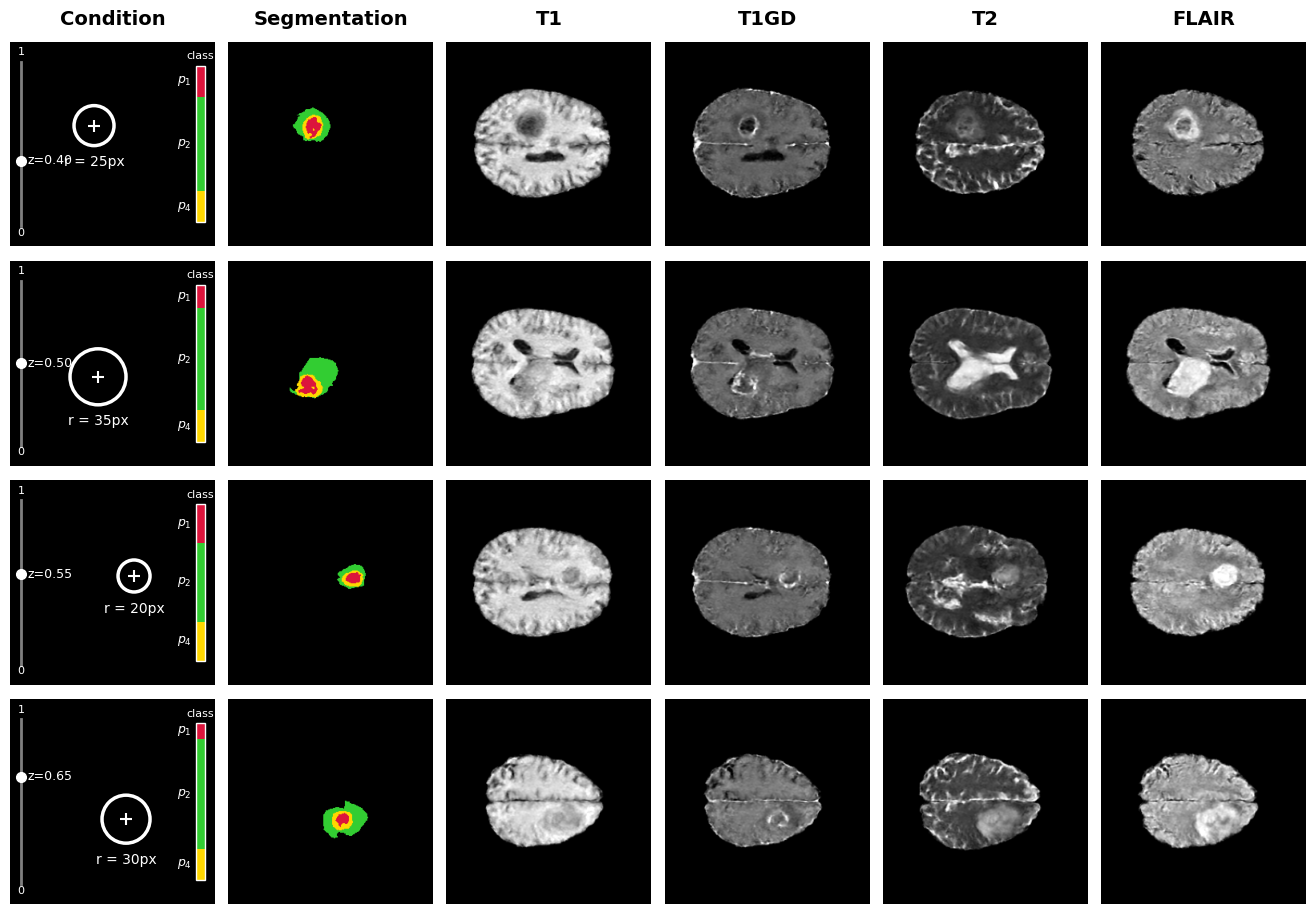}
    \caption{Example synthetic segmentation-MRI pairs. Each row shows a different requested tumour condition, the synthetic tumour segmentation map, and the generated T1, T1Gd, T2 and FLAIR MRI scans corresponding to this map.}
    \label{fig:hero}
\end{figure}

%TODO: discuss hero plot
Figure \ref{fig:hero} demonstrates the end-to-end controllability of the pipeline. Generated segmentation maps remain consistent with the requested spatial and compositional conditions, and the corresponding MRI scans preserve modality-specific appearance and consistency with the segmentation maps.

\begin{table}[t]
\centering
\caption{Adherence to requested tumour morphology across classifier-free
guidance scales. Guidance scale $2.0$ was selected on the validation set and
evaluated on the held-out test set.}
\label{tab:guidance_metrics}
\begin{tabular}{@{}ccccc@{}}
\toprule
\textbf{Guid.} &
\shortstack{\textbf{Centroid}\\\textbf{(px)}} &
\shortstack{\textbf{Radius}\\\textbf{(px)}} &
\shortstack{\textbf{Outside}\\\textbf{(\%)}} &
\shortstack{\textbf{Prop.\ MAE}\\\textbf{(pp)}}  \\
\midrule
\multicolumn{5}{l}{\textit{Validation}} \\
1.0 & 1.232 & 1.120 & 2.047 & 1.239 \\
1.5 & \textbf{0.585} & 0.841 & 1.739 & 0.966\\
2.0 & 0.633 & \textbf{0.683} & \textbf{1.574} & 0.820  \\
2.5 & 0.587 & 0.758 & 1.596 & \textbf{0.770} \\
\midrule
\multicolumn{5}{l}{\textit{Held-out test ($s=2.0$)}} \\
2.0 & 0.842 & 0.738 & 1.789 & 0.795 \\
\bottomrule
\end{tabular}
\end{table}

Table \ref{tab:guidance_metrics} evaluates the ability of Seg-LDM to generate tumours that adhere to the requested features. At the selected guidance scale
of $s=2.0$, performance remained consistent between the
validation and held-out test sets. On the test set, generated tumours achieved a mean centroid error of $0.842$ pixels and a tumour-subregion proportion MAE of $0.795$ percentage points, while only $1.789\%$ of tumour pixels fell outside the requested spatial region on average.

\begin{table}[t]
\centering
\small
\caption{FID comparison between the MRI backbone and ControlNet. Per-modality scores use 6,918 samples; the pooled score uses 27,672 samples.}
\label{tab:fid_results}
\begin{tabular}{@{}lccc@{}}
\toprule
\textbf{Evaluation} &
\textbf{Backbone} &
\textbf{ControlNet} &
$\boldsymbol{\Delta}$\textbf{FID} \\
\midrule
T1$_{6918}$    & 54.53 & \textbf{46.52} & $-8.01$ \\
T1Gd$_{6918}$  & 57.19 & \textbf{46.99} & $-10.20$ \\
T2$_{6918}$    & 41.89 & \textbf{31.14} & $-10.75$ \\
FLAIR$_{6918}$ & 56.18 & \textbf{50.41} & $-5.77$ \\
\midrule
Pooled$_{27672}$ & 42.59 & \textbf{36.12} & $-6.47$ \\
\bottomrule
\end{tabular}
\end{table}

\noindent
Image fidelity was evaluated by comparing the same MRI diffusion backbone with the ControlNet branch enabled or disabled. ControlNet used real held-out segmentations.
Table \ref{tab:fid_results} demonstrates that ControlNet reduced Fréchet inception distance (FID) across all four modalities, with the largest improvements for T2 and T1Gd. Pooled FID decreased from $42.59$ to $36.12$. These results assess MRI synthesis conditioned on real tumour masks; they do not quantify the fidelity of the complete pipeline using Seg-LDM-generated masks.

Cross-study FID comparison is limited by differences in datasets, dimensionality, and feature extractors. Akbar et al.\ \cite{akbar2024brain} reported FIDs of roughly 16 to 85 across 2D GAN and diffusion models on BraTS, while the recent label-guided 3D-LLDM \cite{kim20263dlldmlabelguided3dlatent} reported 28.31 on hepatic MRI. We treat these values as contextual rather than direct baselines, and use the identically evaluated MRI-LDM backbone as our primary comparison.

\begin{table}[t]
\centering
\caption{Mean reader quality ratings (1--10) across MRI modalities and image groups.}
\label{tab:reader_study}
\begin{tabular}{lccc}
\toprule
\textbf{Modality} &
\textbf{Real} &
\textbf{ControlNet} &
\textbf{Fully Synthetic} \\
\midrule
T1    & \textbf{6.33} & \textbf{6.33} & 5.58 \\
T1Gd  & 7.08 & \textbf{7.50} & 7.17 \\
T2    & 6.25 & 6.33 & \textbf{7.33} \\
FLAIR & 6.75 & \textbf{7.17} & 7.00 \\
\midrule
Overall & 6.60 & \textbf{6.83} & 6.77 \\
\bottomrule
\end{tabular}
\end{table}

Four readers each rated 36 segmentation--MRI pairs on a 1--10 scale from three groups: real segmentation with real MRI, real segmentation with synthetic MRI (ControlNet), and fully synthetic segmentation--MRI pairs. Mean ratings were 6.60, 6.83, and 6.77 for the real, ControlNet-generated, and fully synthetic groups, respectively. These descriptive results provide an exploratory assessment of perceived quality; the small reader panel and limited image sample do not establish equivalence between real and synthetic images.

\section{Conclusion}

We introduced a two-stage conditional latent diffusion framework for controllable glioblastoma brain MRI synthesis. The conditional Seg-LDM enables explicit control over the spatial location, extent, and composition of brain tumour, which can be supplied to a ControlNet architecture that generates a corresponding multi-modal 2D brain MRI slice consistent with the requested pathology. The promise of this approach is verified by qualitative analysis of the synthetic images, with expert evaluation providing complementary assessment of image realism. Future work will extend this framework into native volumetric generation.
All the code used in this work is available at: \url{https://github.com/SteveG365/GBM-Diff}.

\section{Compliance with Ethical Standards}

The imaging data used in this study were obtained from publicly available, de-identified datasets. The expert reader study involved voluntary assessment of de-identified and synthetic images. In accordance with institutional requirements, formal ethical approval and waiver were not required.

\section{Acknowledgements}

The authors thank Dr. Qinmei Xu, Dr. Henriette Burger, Dr. Armin Gretschel, and Dr. Derek Jackson for their participation in the expert reader study. No funding was received for conducting this study. The authors have no relevant financial or non-financial interests to disclose.

%The authors thank Dr. , Dr. [Name], Dr. [Name], and Dr. [Name] for their participation in the radiological reader study and for providing expert clinical assessments.

%\vfill\pagebreak

\bibliographystyle{IEEEbib}
\bibliography{refs}

@inproceedings{hoddpm,
    author = {Ho, J. and Jain, A. and Abbeel, P.},
    title = {Denoising Diffusion Probabilistic Models},
    booktitle = {Advances in Neural Information Processing Systems},
    pages = {6840--6851},
    volume = {33},
    year = {2020}
}

@inproceedings{rombach2022LDM,
    author    = {Rombach, R. and Blattmann, A. and Lorenz, D. and Esser, P. and Ommer, B.},
    title = {High-Resolution Image Synthesis with Latent Diffusion Models},
    booktitle = {IEEE/CVF Conference on Computer Vision and Pattern Recognition},
    pages     = {10674--10685},
    year      = {2022},
}

@inproceedings{zhang2023ControlNet,
    author    = {Zhang, L. and Rao, A. and Agrawala, M.},
    title     = {Adding Conditional Control to Text-to-Image Diffusion Models},
    booktitle={IEEE/CVF International Conference on Computer Vision},
    pages     = {3813--3824},
    year      = {2023}
}

@inproceedings{dhariwal2021diffusionmodelsbeatgans,
    author = {Dhariwal, P. and Nichol, A.},
    title = {Diffusion Models Beat {GAN}s on Image Synthesis},
    booktitle = {Advances in Neural Information Processing Systems},
    pages = {8780--8794},
    volume = {34},
    year = {2021}
}

@inproceedings{ho2022classifierfreediffusionguidance,
  author    = {Ho, J. and Salimans, T.},
  title     = {Classifier-Free Diffusion Guidance},
  booktitle = {NeurIPS Workshop on Deep Generative Models and Downstream Applications},
  year      = {2021},
}

@inproceedings{ronneberger2015unet,
  author    = {Ronneberger, O. and Fischer, P. and Brox, T.},
  title     = {{U-Net}: Convolutional Networks for Biomedical Image Segmentation},
  booktitle = {Medical Image Computing and Computer-Assisted Intervention},
  volume    = {9351},
  pages     = {234--241},
  year      = {2015},
}

@inproceedings{puyvelde2026CONFLUX,
  author    = {{Van Puyvelde}, M. and Gulluk, H. and {Van Criekinge}, W. and Gevaert, O.},
  title     = {{CONFLUX}: A Latent Diffusion Model for {3D} Chest-{CT}
               Synthesis with {RL} Post-Training},
  booktitle = {Medical Image Computing and Computer Assisted Intervention Workshops and Challenges},
  volume    = {17258},
  year      = {2026}
}

@inproceedings{pinaya2022brainimaginggenerationlatent,
    author={Pinaya, W. and others},
    title={Brain Imaging Generation with Latent Diffusion Models},
    booktitle={Deep Generative Models},
    year={2022},
    pages={117--126},
    volume = {13609}
}

@article{khader2023medicaldiffusiondenoisingdiffusion,
  author  = {Khader, F. and others},
  title   = {Denoising Diffusion Probabilistic Models for {3D} Medical Image Generation},
  journal = {Scientific Reports},
  volume  = {13},
  number  = {1},
  pages   = {7303},
  year    = {2023},
}

@article{dorjsembe2024conditional,
  author   = {Dorjsembe, Z. and Pao, H. and Odonchimed, S. and Xiao, F.},
  title    = {Conditional Diffusion Models for Semantic {3D} Brain {MRI} Synthesis},
  journal  = {IEEE Journal of Biomedical and Health Informatics},
  year     = {2024},
  volume   = {28},
  number   = {7},
  pages    = {4084--4093}
}

@article{calabreseUCSFGlioblastoma2022,
  title = {The {University} of {California San Francisco Preoperative Diffuse Glioma} {{MRI}} {Dataset}},
  author = {Calabrese, E. and others},
  year = 2022,
  journal = {Radiology: Artificial Intelligence},
  volume = {4},
  number = {6},
  pages = {e220058}
}

@article{bakasUPENNGlioblastoma2022,
  title = {The {{University}} of {{Pennsylvania}} Glioblastoma ({{UPenn-GBM}}) Cohort: Advanced {{MRI}}, Clinical, Genomics, \& Radiomics},
  author = {Bakas, S. and others},
  year = 2022,
  journal = {Scientific Data},
  volume = {9},
  number = {1},
  pages = {453}
}

@inproceedings{zhang2024lefusion,
  author    = {Zhang, H. and others},
  title     = {{LeFusion}: Controllable Pathology Synthesis via Lesion-Focused Diffusion Models},
  booktitle = {International Conference on Learning Representations},
  year      = {2025}
}

@article{KEBAILI2025MultiModal,
    author = {Kebaili, A. and Lapuyade-Lahorgue, J. and Vera, P. and Ruan, S.},
    title = {Multi-modal {MRI} synthesis with conditional latent diffusion models for data augmentation in tumor segmentation},
    journal = {Computerized Medical Imaging and Graphics},
    volume = {123},
    pages = {102532},
    year = {2025}
}

@inproceedings{kim20263dlldmlabelguided3dlatent,
  author    = {Kim, K. and others},
  title     = {{3D-LLDM}: Label-Guided {3D} Latent Diffusion Model for
               Improving High-Resolution Synthetic {MR} Imaging in
               Hepatic Structure Segmentation},
  booktitle = {IEEE International Symposium on Biomedical Imaging},
  year      = {2026}
}

@misc{kofler2025brainles,
      title={BrainLesion Suite: A Flexible and User-Friendly Framework for Modular Brain Lesion Image Analysis}, 
      author={Kofler, F. and others},
      year={2025},
      howpublished = {arXiv preprint arXiv:2507.09036}
}

@article{akbar2024brain,
  author  = {Akbar, M. U. and Larsson, M. and Blystad, I. and Eklund, A.},
  title   = {Brain Tumor Segmentation Using Synthetic {MR} Images -- A Comparison of {GANs} and Diffusion Models},
  journal = {Scientific Data},
  volume  = {11},
  number  = {1},
  pages   = {259},
  year    = {2024}
}

\end{document}